\pdfoutput=1
\documentclass[11pt]{article}

\usepackage[]{EMNLP2022}

\usepackage{times}
\usepackage{latexsym}
\usepackage[linesnumbered,ruled,vlined]{algorithm2e}
\SetAlFnt{\small}
\SetAlCapFnt{\small}
\SetAlCapNameFnt{\small}
\usepackage{algorithmic}
\algsetup{linenosize=\tiny}
\usepackage{tablefootnote}

\usepackage{listings}
\lstdefinelanguage{dict}{
    breaklines=true,
    breakatwhitespace=true,
    basicstyle=\ttfamily\small,
    upquote=true,
    breakindent=0pt,
}

\usepackage[OT2, T1]{fontenc}
\usepackage[russian, english]{babel}
\usepackage{CJKutf8}

\newcommand{\gd}[1]{\textcolor{blue}{#1}}
\newcommand{\mn}[1]{\textcolor{brown}{#1}}

\usepackage[utf8]{inputenc}

\usepackage{microtype}

\usepackage{graphicx}
\usepackage{amssymb}
\newcommand{\mf}[1]{\textcolor{orange}{}}
\usepackage{verbatim}
\usepackage{multirow}
\title{A baseline revisited: Pushing the limits of multi-segment models for context-aware translation}
\author{Suvodeep Majumder$^*$ \\
  NCSU, Amazon \\
  \texttt{smajumd3@ncsu.edu} \\\And
  Stanislas Lauly$^*$ \\
  Amazon \\
  \texttt{laulysl@amazon.com} \\\And
  Maria Nadejde \\
  Amazon \\
  \texttt{mnnadejd@amazon.com} \\\AND
  Marcello Federico \\
  Amazon \\
  \texttt{marcfede@amazon.com} \\\And
  Georgiana Dinu \\
  Amazon \\
  \texttt{gddinu@amazon.com}}

\begin{document}
\maketitle
\begin{abstract}
This paper addresses the task of contextual translation using multi-segment models. Specifically we show that increasing model capacity further pushes the limits of this approach and that deeper models are more suited to capture context dependencies. Furthermore, improvements observed with larger models can be transferred to smaller models using knowledge distillation. Our experiments show that this approach achieves competitive performance across several languages and benchmarks, without additional language-specific tuning and task specific architectures.

%In this paper we investigate how to improve contextual translation models by using already existing tools, like vanilla Transformer model and without relying on new architectures. We first observe the relation between increasing capacity and contextual translation performance, using wider vs deeper models, and reveal that deeper models are more suited to capture such dependencies. We then show that long term dependencies learned by a large model can be transferred to a smaller one using knowledge distillation. 
\end{abstract}
\def\thefootnote{*}\footnotetext{These authors contributed equally to this work}\def\thefootnote{\arabic{footnote}}
\section{Introduction}

\label{sec:intro}

% \gd{TODO:add: \cite{DBLP:journals/corr/abs-2105-06977} \cite{DBLP:journals/corr/abs-2005-03393}}

The quality of NMT (Neural Machine Translation) models has been improving over the years and is narrowing the gap to human translation performance \cite{DBLP:journals/corr/abs-1803-05567}. Until recently, most of the MT research has focused on translating and evaluating sentences in isolation, ignoring the context in which these sentences occur. Simplifying the translation task this way has its advantages: data sets are easier to create, models are computationally more efficient and human evaluations are faster\footnote{With full document context, annotation time per task increases by 68\% according to \citet{grundkiewicz-etal-2021-user}.}. 

%motivate contextual formulation
While initial work failed to show significant differences in standard metrics \citep{tiedemann-scherrer-2017-neural}, the impact of ignoring context has been investigated more closely in recent years \cite{DBLP:journals/corr/abs-2105-06977}. Targeted testing has shown poor performance on discourse-related phenomena \cite{mueller2018,bawden-etal-2018-evaluating, voita-etal-2019-context, jwalapuram2020contextaware, maruf, DBLP:journals/corr/abs-2005-03393} (see  Table~\ref{tab:targ_testsets2} for examples). Furthermore, without context, human evaluation fails to expose all translation errors and leads to rush conclusions on achieving human parity \cite{laubli-etal-2018-machine}. It is thus important to start addressing the MT task in a formulation that is closer to its true complexity and bridges the gap to the real communication needs of the users.

%It is thus becoming clear that the gap to human level performance cannot be accurately measured, or bridged, without paragraph- or document-level translation to address a variety of discourse phenomena, such as lexical cohesion, translation of pronouns, ellipsis, or consistency in formality, style or verb forms. Examples of such discourse phenomena are included in Table~\ref{tab:targ_testsets_appendix}.

%\footnote{More recently \cite{DBLP:journals/corr/abs-2104-14478} show that the human parity gap is measurable in both sentence and document level translation under a fine grained error category annotation.}. Content such as conversational data, subtitles, user-generated text, and other content where sentence boundaries lead to short ambiguous input, with ambiguities that can't be resolved without context. 
%Figure~\ref{tab:targ_testsets2} exemplifies sentence-level translation errors from the test sets use.

This paper tackles the problem of context-aware translation by re-visiting a straightforward multi-sentence translation approach which is considered a baseline in the literature. Our comprehensive experiments show that by leveraging deeper transformer models in combination with knowledge distillation methods, this baseline leads to an effective and robust alternative to specialized architectures proposed in the literature. The paper's contributions are: %~\cite{, DBLP:journals/corr/abs-2005-03393}. T %Our results are based on four language pairs EN-DE, EN-FR, EN-RU and ZH-EN, and leads to the following contributions:

\begin{itemize}
    \item We show that multi-sentence translation can benefit from increased-capacity transformer models and that deeper models are better at learning contextual dependencies than wider models. 
    %\item Next, we show, with fixed data conditions and model capacity, not all large capacity models with different configuration have equal understanding of contextual dependencies. 
    % % @suvodm - should I add that ? next sent
    % \item NOT CLEAR Our training approach of contextual models with equal number of with and without context segments alleviates the BLEU score drop issue for contextual models as reported in literature.
    %\item \gd{its not clear what we alleviate here} We show that adding single sentences to the data used in training we alleviates quality issues reported in the literature.
    \item We further show that distilled models can learn contextual dependencies from larger models, while reducing computational cost and increasing robustness to input length variations.
    \item Finally, results on four language pairs confirm that the approach achieves high performance for both contextual and \textit{single-segment} translation tasks. %This strengths the validity of our findings.
\end{itemize}

\section{Multi-segment translation models}
\label{sec:Multi-segment translation models}

Throughout this paper, we implement context-aware translation models as multi-segment models, as initially proposed in \citet{tiedemann-scherrer-2017-neural} and further used in \citet{fernandes21acl, lopes-etal-2020-document} among others. 

\begin{table*}[]
\centering
{\small{
\begin{tabular}{ll}
\textbf{Input} & \textbf{Output} \\
\hline
 \textless start \textgreater Fire? \textless sep \textgreater Well, put it out, why don't you? \textless end \textgreater & \textless start \textgreater Ein Feuer? \textless sep \textgreater \textbf{Na dann löscht er doch!} \textless end \textgreater \\
  \textless start \textgreater Well, put it out, why don't you? \textless end \textgreater & \textless start \textgreater \textbf{Na dann löscht er doch!} \textless end \textgreater \\
\end{tabular}
}}
\caption{Parallel training data contains both segments in isolation as well as concatenated segments. Example is demonstrative, from the EN-DE anaphora test set \cite{mueller2018}. At inference time, only the translations of target segments (in bold) are used.}
\label{tab:Data format}
\end{table*}

% Specifically, we use document-level parallel data which is transformed to contain concatenated, multi-segment input. In this paper, we concatenate two consecutive segments, both source and target, to create a new data point. As we aim to create single translation models which can perform both translation in-context and in isolation, we duplicate the training data to contain both. Pseudo-code is given in Algorithm \ref{algo:1}.

\paragraph{Multi-segment data points} We use document-level parallel data which is transformed to contain concatenated, multi-segment input. Specifically, we restrict this work to two consecutive sentences. The source and target sides are concatenated using a special delimiter token and added to the training. While not strictly a requirement, the special token allows the extraction of the context-aware translation for the second, target sentence. Prior context-aware architectures can be categorized with respect to the use of context as using: source-side, target-side or both. As it generates both sentence translations jointly, the multi-segment approach takes advantage of both source- and target-side context at train-time. However, it does not use the context reference translation during inference and multi-segment input is simply translated as a continuous output sequence.

\paragraph{Training data} We aim to create single translation models which can perform both translation in-context and in isolation. For this reason, we start from a training set including context for each parallel sentence and create a duplicate of it by removing the context information.  All the contextual models (Ctx) are trained on this joint single- and multi-segment data, while the sentence-level baselines (Bl) use only single sentences. Note that although the data size varies between Bl and Ctx models, the data is effectively identical and all the models are trained using the same stopping criteria, thus conferring no special advantage to any of the models. Table \ref{tab:Data format} exemplifies the training data.

\section{Experimental setup}
\label{sec:experiments}

% \gd{Set up the rest of the paper} 

We perform experiments in four language arcs, English to German (EN-DE), English to French (EN-FR), English to Russian (EN-RU) and Chinese to English (ZH-EN).

\subsection{Training}

We use the WMT2019 data set for EN-DE, Open Subtitles 2018 for EN-FR and EN-RU and UN Parallel Corpus V1.0 for ZH-EN, all four containing document-level data. The data sets vary in size from 4M segments for EN-DE to 17.4M for ZH-EN (see Appendix~\ref{sec:app:sockeye_arguments}, Table~\ref{tab:Datasets} for details). Development data consists of the News task 2019 development set for DE, IWSLT 2019 for FR and newstest2019 for RU and ZH respectively. In all conditions the development data mirrors the training data, meaning that it is duplicated to contain both multi- and single segments data for contextual models, and original and distilled data for distillation experiments. In preliminary experiments we found this to play an important role.

Models use the Transformer architecture~\cite{DBLP:journals/corr/VaswaniSPUJGKP17}. We start with a baseline architecture of 6:2 encoder:decoder layers and 2048 feed-forward width, which subsequent experiments increase in decoder depth and feed-forward width respectively. Training is done with Sockeye \cite{domhan-etal-2020-sockeye}. See Appendix \ref{sec:app:sockeye_arguments} for a complete list of training parameters. 

\subsection{Testing} We measure performance of contextual models using both targeted and non-targeted testing.

\textbf{Non-targeted tests} consists of contextual, document-level data which is \textit{not} selected to focus on discourse phenomena. For EN-DE we use the test set splits made available in \cite{maruf-etal-2019-selective}: TED (2.3k segments), News-Commentary (3k) and Europarl (5.1k). We use  IWSLT15 (1k)~\citep{cettolo-etal-2012-wit3} for EN-FR, WMT newstest2020 (4k) ~\citep{barrault-etal-2020-findings} for EN-RU and finally WMT newstest2020 (2k) ~\citep{barrault-etal-2020-findings} for ZH-EN. While contextual models may improve performance on these data sets, previous work suggests that the effects are minimal in high-resources scenarios with strong sentence-level baselines~\citep{lopes-etal-2020-document}. 
%For this reason, we use this data as generic testing, to ensure that multi-segment models do not degrade translation performance. 
%WMT newstest2019 (2k)~\citep{barrault-etal-2019-findings} for EN-DE

% \textbf{Targeted tests} have been developed in order to evaluate performance on discourse phenomena. \gd{better segway} 
% We measure accuracy on translation of discourse phenomena using contrastive translation pairs from targeted testsets described in Table~\ref{tab:targ_testsets}. A contrastive translation pair consists of a correct human-generated reference translation and a variant of it where a pronoun, or another linguistic unit to be studied, is swapped with an incorrect one. To complement accuracy of contrastive evaluations, we also use targeted test sets and their references to measure standard translation metrics. Table~\ref{tab:targ_testsets2} shows examples from these data sets. %Examples for other targeted test sets can be found in Appendix~\ref{sec:app:sockeye_arguments}, Table~\ref{tab:targ_testsets_appendix}.

\textbf{Targeted tests} have been developed in order to evaluate performance on discourse phenomena. Table~\ref{tab:targ_testsets} lists the test sets used in this paper. \footnote{While highly relevant, data created by \citep{yin2021does} has not been released at the time of writing this paper.} These test sets contain \textit{contrastive} translation pairs, consisting of a correct human-generated translation, and a variant of it where a pronoun, or another linguistic unit of interest, is swapped with an incorrect one. Table~\ref{tab:targ_testsets2} shows examples from these data sets. 

To complement accuracy of contrastive evaluations, we also use targeted test sets and their references to measure standard translation metrics. 

%Contrastive pairs allow us to isolate a linguistic phenomena of interest, such as the translation of anaphoric pronouns. 

% updated and moved the table to appendix

%TODO: Add the new lang test set details
\begin{table}[t]
{\small{
\centering
\begin{tabular}{llll}
\hline
\textbf{LP}  & \textbf{Type} & \textbf{Size} &  \textbf{Source} \\
\hline
EN-DE & Anaphora           & 12,000 &  \cite{mueller2018}\\
\hline
%EN-FR & Anaphora           & 300    &  \cite{bawden-etal-2018-evaluating} \\
EN-FR & Anaphora     & 12,000    &  \cite{lopes-etal-2020-document} \\
\hline
EN-RU & Deixis             & 3,000 &  \cite{voita-etal-2019-good} \\ 
     & Lex-coh             & 2,000 &  \\ 
     & Ellipsis-vp          & 500 &  \\ 
     & Ellipsis-infl  & 500 &  \\ \hline
ZH-EN & Anaphora     & 500    &  \cite{jwalapuram2019evaluating} \\
\end{tabular}
\caption{Targeted test sets used for evaluating discourse phenomena.}
\label{tab:targ_testsets}
}}
\end{table}
%83 73 86

\begin{table}[t]
\small
\centering
{\resizebox{\linewidth}{!}{
\begin{tabular}{lll}
\hline
DE & Src & I forgot to confide it to you.\\
& Ctx & What's your plan?\\
& Ctx-tgt & Was hast du vor? \\
& Ref & Ich vergaß, \textbf{es} euch zu vertraun.\\
& Contr & Ich vergaß, \textbf{sie} euch zu vertraun.\\
\hline
 FR & Src & And where's it coming from?\\
 & Ctx & A sort of mist.\\
 & Ctx-tgt & Une sorte de brume.\\
 & Ref & Et \textbf{elle} vient d'où ? \\
 & Contr & Et \textbf{il} vient d'où ? \\
 \hline
 RU  & Src & Identity theft. \\
  & Ctx & And I solved another crime.\\
 & Ctx-tgt & \foreignlanguage{russian}{И этим решил еще одно преступление.}\\
 & Ref & {\foreignlanguage{russian}{\textbf{Кражу.}}}\\
 & Contr & {\foreignlanguage{russian}{\textbf{Кража.}}} \\
 \hline
 ZH  & Src & \begin{CJK}{UTF8}{gbsn}情况就是这样\end{CJK} \\
  & Ctx & \begin{CJK}{UTF8}{gbsn}斐济人就好像生来就是打 7人\end{CJK} \\
  & & \begin{CJK}{UTF8}{gbsn}制橄榄球的, 而英国队仍是初出茅庐\end{CJK} \\
 & Ctx-tgt & \textbf{It} was as if Fiji had been born to play 7s, \\
  & & while GB are still learning the trade .\\
 & Ref & \textbf{Which} is pretty much how it is.\\
 & Contr & That is pretty much how it is. \\
 \hline
\end{tabular}
}}
\caption{Targeted test set examples. Models are assessed as correct if they score the reference (Ref) higher than a contrastive variant (Contr), given a source segment and its context.}
\label{tab:targ_testsets2}
\end{table}

% @suvodm - add why ZH is bad - only which and that

%\gd{good for summary https://lena-voita.github.io/posts/acl19_context.html}

%As the difference between wrong and correct translations may be reflected only in minor changes at string level, additional metrics have been developed in order to complement quality measures such as BLEU which are not sensitive enough to capture such differences. 

\section{Context-aware translation results}
\label{sec:results1}

% We start by testing the performance of concatenated, multi-segment models on contextual translation benchmarks with another start of the art contextual translation model with context aware decoder called \textbf{CADec}. Next, we use EN-DE as a test bed to further test increasing model capacity. Finally, we report results using the best setting as determined on DE, on FR, RU and ZH, in order to test the robustness of our results.

We begin our experiments by confirming that concatenated models are indeed able to model context dependencies (Section \ref{subsection:results1}). We follow by testing the hypothesis that larger models are better suited for learning the more complex contextual training data (Section \ref{subsection:results2}). In order to avoid over-fitting, we use EN-DE as a development language and subsequently test identical settings on FR, RU and ZH in Section \ref{subsection:results3}.

\subsection{Multi-segment models}
\label{subsection:results1}
For all four language arcs, \textbf{Ctx} models use 6 encoder layers and 2 decoder layers (44M parameters) and are trained using both segments in isolation as well as concatenated context segments. In inference, DE, FR and ZH models use one preceding context sentence, matching the training. However, over 60\% of the targeted RU data exhibits longer dependencies, of up to 3 previous segments. For this reason, targeted EN-RU testing concatenates all three context sentences. Baseline (\textbf{Bl}) models use the same original train data and model architecture, this time trained and used to translate one segment at a time.

% @suvodm - we wanted to use shalow model to be able to increase decoder size - also for practical reason for shalow decoder is to able to decode faster 
%As a baseline (\textbf{Bl}) we use the same original train data and same model architecture, this time trained and used to translate one segment at time. Note, both Bi and Ctx model uses 2 decoder layer configuration, this is because it will allow us to increase the number of decoders for creating larger models for investigating the effect of large capacity models. Also this shallower model configuration will allow faster decoding which is desirable for practical purposes.

Results are shown in Table \ref{tab:results0}. As observed by previous work, concatenated models are considerably better than their context-ignoring counterparts, particularly on targeted test sets. In contrastive testing, accuracy increases by 20-30\% in absolute values in all languages, with the exception of the lexical cohesion data set in RU and anaphora data set in ZH.

% @suvodm - added text saying our approach improved Ctx model - it reduces the BLEU score drop issue.

For \textit{non-targeted} testing, {Ctx} models significantly out-perform the {Bl} models in 4 out of the 6 test sets. This differs from previous work, where contextual models using the concatenation approach are reported to degrade BLEU scores: \citet{tiedemann-scherrer-2017-neural} measure 0.6 BLEU drop, \citet{voita-etal-2019-good} show a 0.84 drop for RU, \citet{lopes-etal-2020-document}, 1.2, and \citet{junczys2019microsoft} shows a BLEU degradation of 1.5. These results indicate that our approach to train the contextual model with both contextual and non-contextual data alleviates the issue of quality degradation. %BLEU score drop and further leads BLEU point gains except for FR, where we see a slight drop in BLEU score~\footnote{A related observation was made in \cite{lopes-etal-2020-document} for FR models.}.

\begin{table}[]
\centering
{\small{
\begin{tabular}{lllc|cc}
Arc &	Metric&	Test set&	Targeted &	Bl&	Ctx \\
\hline
DE&	BLEU&	TED&	&	19.9&	\textbf{22.4}\\
	&	& News &	&	26.1&	\textbf{29.5}\\
	&	& Europarl &	&	29.3&	\textbf{31.5} \\
	%&	&WMT19&	&	29.9&	\textbf{30.1}\\
	&BLEU&	ContraPro&	\checkmark &	20.1&	\textbf{21.1}\\
	&Acc&	ContraPro&	\checkmark&	0.50&	\textbf{0.70}\\
\hline
FR&	BLEU&	IWSLT&	 & \textbf{40.0} & 39.7 \\		
	&BLEU& LCPT\tablefootnote{\href{https://github.com/rbawden/Large-contrastive-pronoun-testset-EN-FR}{Large-contrastive-pronoun-testset-EN-FR} (LCPT)} &		\checkmark	 & 27.9 & \textbf{32.5} \\
	&Acc& LCPT &		\checkmark	& 0.74 & \textbf{0.87}\\
	&Acc& Anaphora &		\checkmark	& 0.50 & \textbf{0.72}\\	
\hline
RU&	BLEU&	WMT20 &			& 13.6 & \textbf{14.6}\\	
	&Acc& Deixis &	\checkmark		& 0.50 & \textbf{0.83}\\
    &Acc& Lex-coh &	\checkmark		& 0.46& \textbf{0.47}\\	
	&Acc& Ellipsis-vp &	\checkmark	& 0.20 & \textbf{0.60}\\
	&Acc& Ellipsis-infl &	\checkmark	& 0.52 & \textbf{0.68}\\	
\hline
ZH&	BLEU&	WMT20 &			& 21.2 & \textbf{21.4}\\	
	&Acc& Eval-anaphora &	\checkmark		& 0.58 & \textbf{0.61}\\
\end{tabular}
}}
\caption{Concatenated models (Ctx) vs baseline models (Bl) of the same capacity. While all test sets have context, some are targeted towards discourse phenomena, marked as Targeted (see Section \ref{sec:experiments} for details).}
\label{tab:results0}
\end{table}

\subsection{Increasing model capacity}
\label{subsection:results2}
As the multi-segment models are trained on data exhibiting longer dependencies, we investigate the hypothesis that increased model capacity is needed to learn the more complex data distribution. 

Starting with the baseline model used in the previous experiments (\textbf{Ctx}), we investigate two ways to increase its capacity: increasing the depth of the decoder or increasing the width of the feed-forward layers. We test four increased model capacities, by incrementally adding 2 decoder layers to the base model (deep models). For each deep model, we also create an equivalent wide model containing the same number of parameters. This leads to number of parameters ranging from 44M -- the baseline setting -- to 76M. We leave all other settings un-changed. Table \ref{tab:ModelCapacityAppendix} in the Appendix details these architectures. 

\begin{figure*}
\begin{center}
\includegraphics[width=1.02\textwidth]{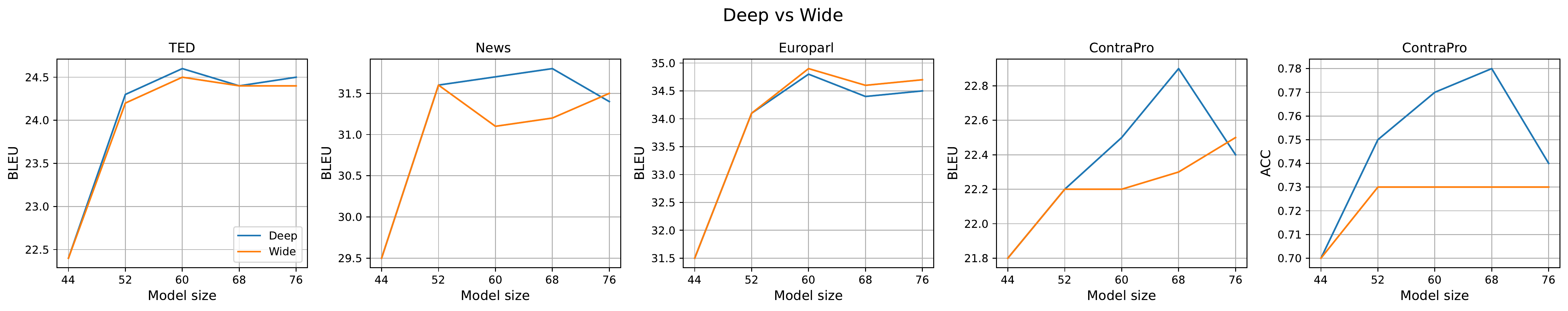}
%{Figures/deep-vs-wide-bleu-all-test-sets.pdf}
\end{center}
\caption{EN-DE, Ctx models when increasing model capacity, measured in millions of parameters: BLEU scores on three non-targeted test sets (Ted, News and Europarl) and targeted metrics (BLEU and Accuracy) on ContraPro.} % \mn{To show only the average use Figures/deep-vs-wide-bleu-avg.pdf}}
\label{fig:results1}
\end{figure*}

Non-targeted and targeted testing results are shown in Figure \ref{fig:results1}. Results show that larger capacity models deliver increased performance across the board. In both cases most of the performance gain comes from the 52M capacity model: +2.2 BLEU gain in non-targeted and +0.4 BLEU/+5\% Accuracy on the ContraPro pronoun test. However targeted metrics show subsequent improvements with increased depth, to +8\% absolute accuracy gain with the 6:8 encoder:decoder configuration. While deeper and wider models perform similarly in non-targeted testing, deeper models are clearly superior on the pronoun translation task: Wider models improve accuracy from 70\% to a maximum of 73\% while deep models achieve 78\%. 

\begin{table*}[]
\centering
{\small{
\begin{tabular}{llcc|cccc}
\multicolumn{4}{c|}{Testing configuration} & \multicolumn{4}{|c}{Model} \\
\hline
Metric & Test set & Targeted & Ctx used &	Bl & Ctx &	Ctx-Deep &	Ctx-Wide\\
\hline
    BLEU & TED & &  &19.9& 22.1 &	24.3&	24.3\\
         & TED & & 	\checkmark          & -   & \textbf{22.4} &  \textbf{24.4}&	\textbf{24.4}\\
         & News & & &26.1& \textbf{29.7} &	31.3&	31.3\\
         & News & & \checkmark	      &  -  & {29.5} &  \textbf{31.8}&	\textbf{31.5}\\
         & Europarl & &  &29.3& 31.5 &	34.4 &	34.5\\  
         & Europarl & & \checkmark	      &  -  & \textbf{31.5} &  \textbf{34.4}&	\textbf{34.7}\\
\hline
BLEU & ContraPro &  \checkmark & 	          &  20.1   & 19.1&  21.2    &20.6	\\
     & ContraPro &  \checkmark & \checkmark  & - & \textbf{20.4}&	 \textbf{22.9}&	\textbf{22.3}\\
Acc  & ContraPro &  \checkmark & 	          &  0.49   & 0.50&  0.51&	0.50\\
     & ContraPro &  \checkmark & \checkmark  & - & \textbf{0.70} &	 \textbf{0.78}&	\textbf{0.73}\\
\end{tabular}
}}
\caption{EN-DE, Bl and Ctx models of standard capacity (44M), and the best Deep and Wide models (68M/76M respectively). All used with and without context at test time.}
\label{tab:results4}
\end{table*}

%\paragraph{Translation in isolation}
As noted in Section \ref{subsection:results1}, Ctx models do not perform worse than sentence-level baselines on any of the contextual data sets, most likely due to the joint single- and multi-segment training regime. Next we investigate if this training leads to ``multi-task" models, that maintain baseline performance also when used \textit{without context}. These experiments use the previously trained models, and contextual models are tested with/without context at inference time. We contrast baseline single-segment models of standard capacity (Bl), similar multi-segment models (Ctx), and the best Deep and Wide models as previously determined (capacity 68M and 76M respectively). 

% Results are shown in Table \ref{tab:results4}. As it can be observed, contrary to prior studies, translation with context for our contextual (Ctx) models significantly out-performs translation without context for baseline (Bi) model. Prior studies for contextual model using concatenation approach shows BLEU score drop between baseline and contextual model, i.e., \citet{tiedemann-scherrer-2017-neural} shows BLEU score drop of 0.6, \citet{voita-etal-2019-good} reports BLEU score drop of 0.84, for \citet{lopes-etal-2020-document} the drop is 1.2 and \citet{junczys2019microsoft} shows a BLEU score drop of 1.5. We see that using our approach to train the contextual model with both contextual and non-contextual data alleviates the issue of BLEU score drop and further leads to +2/+3 BLEU point gains. Note that the standard capacity \textbf{Bl} and \textbf{Ctx} models are used identically at test-time and only differ in the use of training data: Ctx duplicates the training data by concatenating two adjacent segments.

% We also observe, in \textit{non-targeted} testing, multi-segment models used without context approach the optimal performance. 

Results are shown in Table \ref{tab:results4}. Interestingly, in \textit{non-targeted} testing, multi-segment models used \textit{without} context approach the optimal performance. Therefore the improvements due to the use of context are smaller than indicated by Section \ref{subsection:results1}: +0.5 and +0.2 BLEU in News and negligible or non-existent in the other domains.~\footnote{A related observation was made in \cite{lopes-etal-2020-document} where it is shown that if strong baselines are used, no contextual model tested brings any improvements over the context-ignoring baselines in IWSLT sets for De and Fr.} Specifically, standard capacity {Ctx} models outperform {Bl} ones by +2/+3 BLEU points when used {without} context. Note that these only differ in the use of training data: Ctx duplicates the training data by concatenating two adjacent segments. %This alone leads to +2/+3 BLEU point gains. %In future work we plan to further investigate if a) adding multi-segment training data leads to reliable single-segment translation quality improvements and b) multi-segment models can be improved when used as contextual models, by reducing the error propagation in translating long inputs.

In \textit{targeted} tests, the multi-segment models that ignore context do not outperform the baselines. This confirms that ContraPro is a good benchmark for isolating context-aware phenomena, as the task cannot be solved with a strong baseline.

\subsection{Results on FR, RU and ZH translation}
\label{subsection:results3}
In this section we investigate whether EN-DE results carry over to the FR, RU and ZH translation tasks, by testing the optimal DE configurations without any additional language- or task-specific tuning. We test the baseline model (single-segment, standard capacity) against the best contextual model as determined on DE, the multi-segment model using 6 encoder and 8 decoder layers. 

\begin{table}[t]
\centering
{\small{
\begin{tabular}{p{.2cm}lp{1.9cm}p{0.4cm}c|cp{0.4cm}cc}
% \hline
% & \multicolumn{4}{c|}{Testing configuration} & \multicolumn{2}{|c}{Model} \\
% Arc & Metric & Test set & Targeted & Ctx used &	Bl & Ctx-Deep \\
& \multicolumn{4}{c|}{Testing configuration} & \multicolumn{2}{|c}{Model} \\\hline
Arc & Metric & Test set & Targ-eted & Ctx &	Bl & Ctx-Deep \\
\hline
  FR  & BLEU & IWSLT & & 	      &  40.0  & \textbf{40.8} \\
      &  & IWSLT & & \checkmark & - & 40.0 \\  
      &  BLEU& LCPT & \checkmark & 	  &  27.9      &   31.5  \\
      &  & LCPT & \checkmark & \checkmark  & - & \textbf{32.3} \\
      & Acc & LCPT &  \checkmark & 	          & 0.74    & 0.79 \\
      &  & LCPT & \checkmark & \checkmark  & - & \textbf{0.90} \\
      \hline
RU & BLEU & WMT20 &   & 	          &   13.6  & \textbf{18.9}\\
      &       & WMT20 &   & \checkmark  & - & 16.5 \\
      & Acc   & deixis &  \checkmark & 	          & 0.50  & 0.51 \\
      &       & deixis &  \checkmark & \checkmark  & - & \textbf{0.85} \\      
      & Acc   & lex-coh &  \checkmark & 	          &  0.46   & 0.46 \\
      &       & lex-coh &  \checkmark & \checkmark  & - & \textbf{0.48 }\\      
      & Acc   & ellipsis-vp &  \checkmark & 	          &  0.20  & 0.25 \\
      &       & ellipsis-vp &  \checkmark & \checkmark  & - & \textbf{0.73} \\    
      & Acc   & ellipsis-infl &  \checkmark & 	          &  0.52   & 0.55 \\
      &       & ellipsis-infl &  \checkmark & \checkmark  & - & \textbf{0.80} \\   
      \hline
ZH & BLEU & WMT20 &   & 	          &   21.2 & 22.1	\\
        &       & WMT20 &   & \checkmark  & - & \textbf{22.4} \\
        & Acc   & Eval-anaphora &  \checkmark & 	          & 0.58  & 0.60 \\
        &       & Eval-anaphora &  \checkmark & \checkmark  & - & \textbf{0.62} \\
\end{tabular}
}}
\caption{Single segment models of standard 44M capacity (Bl) and 68M deep multi-segment models (Ctx-Deep). Best results on each task in bold font.}
\label{tab:results5}
\end{table}

% % @suvodm - add cite - add cite saying others also see same effect - also add why RU have worst effect with and without context.
% The DE results carry over to FR, RU and ZH to a large extent: On \textit{non-targeted} testing the best performance is obtained by the deep multi-segment models used \textit{without} context.This again indicates that either the testing fails to capture context dependencies (through content or metric), or that the models implemented are not able to effectively use this context. 
% In targeted testing, deep multi-segment models using context improve performance by a large margin, with the exception of the RU lexical cohesion test set, where the improvements are minimal. Similarly for ZH-EN arc the targeted test set for anaphoric pronoun translation shows minimal improvement, which is consistent with the prior studies~\cite{jwalapuram2020pronoun, jwalapuram2019evaluating} with the same anaphoric pronoun test set. 

% @suvodm - add cite - add cite saying others also see same effect - also add why RU have worst effect with and without context.
Table \ref{tab:results5} shows the results. The DE results carry over to FR, RU and ZH to a large extent. Except for ZH, on \textit{non-targeted} testing the best performance is obtained by the deep multi-segment models used \textit{without} context. On further analysis of the non-targeted EN-RU test set, where the drop is of over 2 BLEU points, we observed that the length divergence between training and test segments is significant; the quality drops dramatically when the segment length deviates by more than 4 standard deviation(sd). (See Appendix~\ref{sec:app:results} for a detailed analysis of segment length variation.).

In targeted testing, deep multi-segment models using context improve performance by a large margin, with the exception of the RU lexical cohesion test set and ZH anaphoric pronoun translation, where the improvements are minimal, corroborating prior studies~\cite{jwalapuram2020pronoun, jwalapuram2019evaluating}.

\section{Distillation}

%TODO: Add resuls for ctx, ctx-deep, student and baseline for Student Model @suvodm
\begin{table}[t]
\centering
{\small{
\begin{tabular}{l|l|c|ll}
&& \multicolumn{1}{c|}{Non-Trgtd} & \multicolumn{2}{c}{Targeted}  \\
&Model & BLEU & BLEU & Acc \\ 
\hline
EN-DE&Ctx-Deep &\textbf{32.0} & 22.9 & \textbf{0.78} \\
%&St dist & 31.3 & 22.9 & 0.75 \\
&Student & 31.4 & \textbf{23.1} & 0.73 \\
&Ctx &  29.5 & 20.1 & 0.70 \\
%&Bl & ?  & 21.8 & 0.70 \\
%& \cite{maruf-etal-2019-selective}-offline & & & 0.69 \\
%& \cite{maruf-etal-2019-selective}-online  & & & 0.73 \\
%& \cite{lopes-etal-2020-document}\gd{concat?}  & & & 0.71 \\
%& \cite{fernandes21acl} 2to2  & & & 0.66 \\
%& \cite{fernandes21acl} 2to2 + pre  & & & 0.80 \\
\hline
EN-FR&Ctx-Deep & 40.0 & 32.3 & \textbf{0.90} \\
%&St dist & 39.8 & 31.9 & 0.86 \\
&Student & \textbf{40.4} & 32.1 & 0.88 \\
&Ctx & 39.7  & \textbf{32.5} & 0.87 \\
%&Bl &   &  & 0.74 \\
%&\cite{lopes-etal-2020-document} concat & & & 0.83\\
\hline
EN-RU&Ctx-Deep & \textbf{16.5} & - & \textbf{0.85} \\
deixis& Student & 16.2 & - & 0.84 \\ %St dist & 16.3 & - & 0.84 \\
&Ctx & 14.6  & - & 0.83 \\
%&Bl & \gd{?}  &  & 0.5 \\
%&\cite{voita-etal-2019-good} Concat & & & 0.83 \\
%&\cite{voita-etal-2019-context} DocRepair & & & 0.92 \\ 
\hline
EN-RU&Ctx-Deep & - &- & \textbf{0.48} \\
lex-coh&Student & - & - & 0.46 \\ %St dist & - &- & 0.46 \\
&Ctx & - & - & 0.46 \\
%&Bl &   &  &  0.46 \\
%&\cite{voita-etal-2019-good} Concat & & & 0.47 \\
%&\cite{voita-etal-2019-context} DocRepair & & & 0.80 \\ 
\hline
EN-RU&Ctx-Deep & -  & - & \textbf{0.73}\\
ellipsis-vp& Student & - & - & 0.66 \\ %St dist & - & - & 0.65 \\
&Ctx & - & - & 0.60 \\
%Bl &   &  & 0.2 \\
%&\cite{voita-etal-2019-good} Concat & & & 0.77 \\
%&\cite{voita-etal-2019-context} DocRepair & & & 0.75 \\ 
\hline
EN-RU&Ctx-Deep & - & - & \textbf{0.80}\\
ellipsis-infl&Student & - & - & 0.69 \\ %&St dist & - & - & 0.70 \\
&Ctx & - & - & 0.68 \\
\hline
ZH-EN&Ctx-Deep & \textbf{22.3} & - & \textbf{0.62}\\
 &Student & 22.2 & - & 0.59 \\ %&St dist & - & - & 0.70 \\
&Ctx & 21.4 & - & 0.60 \\
%&Bl &   &  & 0.52 \\
%&\cite{voita-etal-2019-good} Concat & & & 0.76 \\
%&\cite{voita-etal-2019-context} DocRepair & & & 0.86 \\ 
\end{tabular}
}}
\caption{Teacher (68M deep multi-segment) and student (44M multi-segment) models. Ctx is a multi-segment model of 44M capacity. Non-targeted test sets are News for DE, IWSLT for FR and WMT for RU.} 
\label{tab:KnowledgeDistillation}
\end{table}

The multi-segment models with increased capacity are computationally less efficient than standard single-segment models, with deeper decoders and longer outputs impacting latency and wider models impacting memory. %Notice that latency can the former can be addressed by reusing the computations performed in translating the context, when processing running text such as documents. Marcello: actually what matters is the decoding output length and it's depth. So if you are reducing the decoder depth with KD you are also reducing latency. 

This section investigates the effectiveness of Knowledge Distillation (KD) in compressing contextual models to the original model capacity. %\gd{somewhere highlight that the student models not only use the standard architecture, are multi-task but also are computationally equivalent.}
We employ sequence-level KD as proposed in \citep{kim-rush-2016-sequence}. Specifically, the Deep-Ctx models are the teachers used to translate the training/development data and students are trained on data containing both references and teacher output, as recommended in \cite{gordon2019explaining} among others. Students with $(6,2)$ layers and 44M parameters are trained with the loss:
%\begin{equation*}
$\small
\sum_{j=1}^{J} \sum_{k=1}^{|V|} 1\{\hat{y}_j = k\} \times \log p(y_j=k|\bf{\hat{y}}_{<j}, \bf{x}, \theta),$
%\end{equation*}
where $\bf{\hat{y}}$ is the teacher prediction, $J$ is the target length and $|V|$ is the size of the vocabulary. %For source target pairs, sequence-KD will replace the target, where for monolingual data we simply create new pairs. 

\subsection{Results}
Results in Table \ref{tab:KnowledgeDistillation} show KD can indeed enhance a smaller model’s performance on discourse phenomena. Overall, the performance of the distilled models lies between that of standard capacity models and that of deep models. 

We observe that, unexpectedly, in several test sets, the student model is better than the teacher. We analyze such a test set, EN-DE WMT19, where teacher and student achieve 30.6/31.1 BLEU respectively. We observed that some of this data is paragraph-level and not sentence-level, again leading to a train-test miss-match. We analyzed the performance on the test set against the input length distribution seen in the training data. Results (Table \ref{tab:results7}) show that the student outperforms the teacher when the input length is above 2 standard deviations from the median length. 

We hypothesize that student translations are more robust to variations and less-context sensitive, due to the simpler data distribution that the student is trained on (\citet{Zhou2020Understanding} indeed show that distilled data is less complex under a complexity metric based on cross-entropy across word alignments). While we leave further exploration to future work, we perform a simple experiment to measure the variation observed in translation when context is used or ignored. We measure this irrespective of quality, as the percentage of translations that change at all when context is used. Table \ref{tab:results7.2} shows that indeed student models are less context-sensitive. This is observed across the targeted test set, ContraPro, where translation changes are expected; the gap is even more pronounced on non-targeted test sets, where context-dependence is not part of the data set design. 

\begin{table}[t]
\centering
{\small{
\begin{tabular}{l|cc}
Len. range & Ctx-Deep & Student\\
\hline
(0,m) &	22.7& \textbf{23.0} \\
(m,m + 2* SD] &	\textbf{32.3}& 31.8 \\
(m+2*SD, m+4 *SD] &	32.2& \textbf{32.7} \\
 (m+4* SD,$+\infty$) &	25.1& \textbf{28.1} \\
\hline
\end{tabular}
}}
\caption{BLEU scores on WMT19 EN-DE test set. The test set is split into four partitions, wrt. the input length. For example, the second bin contains input of length between the median length seen in training and 2 sd.}
\label{tab:results7}
\end{table}

\begin{table}[t]
\centering
{\small{
\begin{tabular}{l|cc}
Test set & Ctx-Deep & Student\\
\hline
Ted & 57.0\% & 47.0\% \\
News & 60.5\% & 49.7\% \\
Europarl &	53.6\%& 42.2\% \\
ContraPro &	65.0\%& 57.5\% \\
\hline
\end{tabular}
}}
\caption{EN-DE, percentage of translations that change with the addition of context (ctx used vs. ignored)}
\label{tab:results7.2}
\end{table}

%\gd{include consistency results we obtained later here}

%This indicates that distillation is a promising future work direction for improving translation quality through robustness. 

%@suvodm - should be separate section
\section{Human evaluation}

%\gd{motivation: sentence to point out that the scoring accuracy metric does not guarantee or measure if the correct pronoun is actually used, as pointed it out by the people who came up with it.}

The scoring accuracy metric used with targeted contrastive references does not measure if the system can actually generate the correct pronoun. In a recent analysis, \citet{vamvas-sennrich-2021-limits} show that scoring human-written contrastive references may lead to false positives, in particular for distilled NMT models which have not been exposed to real-data distribution during training.
% "This problem is especially apparent for distilled NMT models, which perform poorly on humanwritten minimal pairs because they were never exposed to such a distribution during training. While this indicates that distilled NMT models are less robust against improbable contexts, human-crafted minimal pairs also become less useful to predict their unconstrained generative behavior."

To address this limitation, we complement the automatic metric with human evaluation of the EN-DE teacher and student models. We sampled 250 examples from ContraPro, selecting samples where the antecedent is in the previous sentence.  % where the second sentence may contain an ambiguous term (pronoun "it") which cannot be translated correctly out of context.
Translators were shown the two consecutive source sentences and their translations. They were asked to rate the quality of both sentences, the context and target, on a scale of 1 to 6 (with increments of 0.2) and to mark if the anaphoric pronoun "it" was correctly translated in the target sentence. Each translator performed two tasks: the first task was to compare the context-ignoring baseline (Bl) and the Ctx-Deep model; the second task was to compare the same baseline with the student model (Student). With this setup, we grounded the evaluation by showing the baseline translations in both tasks.

The inter-annotator agreement on ranking the baseline and teacher models wrt. generic quality of the target sentence\footnote{We compute the ranking as $sign(score_{ctx} - score_{base})$.} is good at 0.55 Krippendorff’s Alpha~\citep{krippendorffalpha}. On assessing the correctness of pronoun translation, the agreement is very high for the teacher at 0.86 and high for the student 0.66.\footnote{We believe annotator fatigue contributed to the decrease in agreement, as translators first judged the teacher outputs and later the student. For the baseline model, which was judged twice, translators agree with themselves 86\% of the time.}

We report the pronoun translation accuracy and the generic quality scores (averaged across annotators and sentences) for the target sentences in  Table~\ref{tab:human_eval_context_no_context_1}. These results show that the Ctx-Deep is significantly more accurate at translating ambiguous pronouns than the baseline (71.8\% vs 28.1\%) and at the same time achieves better generic quality scores (+8\% relative improvement). The student, which has the same capacity and architecture as the baseline, performs better than the automatic accuracy metric suggested: it retains most of the improved accuracy (61.6\% vs 71.8\%) and quality of the teacher model (+7\% vs +8\% ). 

% \mn{Comment out below because we cite related work at the beginning of the section}
%We note that accuracy of the Ctx-Deep on ContraPro computed by scoring contrastive references is 0.79 while the accuracy as judged by translators is 0.71. This indicates that the automatic scoring overestimates the performance. Among others, it does not account for imperfect translations of the context sentence or robustness of the models to noisy inputs. For example, translators do not mark as correct the translation of pronouns appearing in ambiguous contexts such as “It's printing! It's printing! ” (possible meanings - “The poster is being printed”/“The printer/machine is printing”). However when computing accuracy by scoring the contrastive references, the system may be rewarded on such examples. 

Table~\ref{tab:ExamplesStudent} shows an example of translation where the student model disambiguates the anaphoric pronoun "it" and in addition makes better word choices compared to the baseline:  the two occurrences of the verb "jump" ("springt") are correctly translated in both the context and target sentence.
%In the first example, the student model chooses the correct translation for the adjective "overextended" ("überzogen") referring to the "credit card" mentioned in the context sentence.

\begin{table}[t]
\centering
{\small{
\begin{tabular}{l|ll|ll}
\hline
& \multicolumn{2}{c|}{Targeted accuracy (\%)} & \multicolumn{2}{c}{Quality scores}  \\

Model & $Avg.$ & $\Delta_{bl}$ & $Avg.$ & $\%_{bl}$ \\ \hline
Bl & 28.1$_{\pm2.40}$ & - & 4.43$_{\pm0.01}$	& -\\
Ctx-Deep & 71.8 & 42.0 & 4.81 & 8\%  \\
Student& 61.6 & 31.8 & 4.76 & 7\% 
% Base-Deep-44 taskA: 29.8, 4.44

% Model & Avg & Judge-A & Judge-B  & Avg & Judge-A & Judge-B & Avg parity \\
% \hline
%  Base-Deep-60M  & 0.32&0.33& 0.3 & 4.3 & 4.2 & 4.5 & \multirow{2}{*}{106.8\%} \\
%  Deep-60M  & 0.51 & 0.52 & 0.49 & 4.7 & 4.5 & 4.8 & \\
\end{tabular}
}}
\caption{Human evaluation of EN-DE contextual models versus the (context-ignoring) baseline on ContraPro.}
\label{tab:human_eval_context_no_context_1}
\end{table}

\begin{table*}[t]
\centering
{\small{
\begin{tabular}{p{1cm}p{6cm}p{6cm}p{1cm}}
\hline
 & Context sentence  & Targeted sentence & Avg. scores \\ \hline
 
%  Src & The credit bureau has instructed me to destroy \underline{this card}. &  \textbf{It}'s \textbf{overextended}. & \\
% Bl & Das Kreditbüro hat mich angewiesen, \underline{diese Karte} zu zerstören. &  \textbf{Es} ist \textbf{überdehnt}. & 5.2 / 3.8 \\
% Ctx-Student & Das Kreditbüro hat mich angewiesen, \underline{diese Karte} zu zerstören. &   \textbf{Sie} ist \textbf{überzogen}. & 5.2 / 5.8 \\
% \hline
Src & \underline{This} wounded \underline{bird} is \textbf{jumping} all over the place out there. &  And every time it \textbf{jumps}. \textbf{It} gives us more data that we can use. & \\
Bl & \underline{Dieser} verwundete \underline{Vogel} \textbf{zieht} sich dort über den gesamten Ort hinaus.  & Und jedes Mal, wenn es \textbf{springt}, gibt \textbf{es} mehr Daten, die wir verwenden können. & 4.0 / 4.5 \\
Ctx-Student & \underline{Dieser} verwundete \underline{Vogel} \textbf{springt} dort draußen. & Und jedes Mal, wenn er \textbf{springt}, gibt \textbf{er} uns mehr Daten, die wir verwenden können. &  4.4 / 5.4 \\
\hline

\end{tabular}
}}
\caption{Translation examples where the contextual student model is more accurate and has better generic quality compared to the baseline. We underline the antecedent of the ambiguous pronoun and bold words that are translated correctly by the student.  We report the average quality scores for both the context and target sentences.}
\label{tab:ExamplesStudent}
\end{table*}

%\mn{Source-original vs Target-original evaluation}
%TODO(Maria) - Regression analysis with context flag as fixed effect; breakdown by domain, LPs. (!!!! WOULD GO WELL WITH THIS WORK - relation to larger/deeper models and reduction in Grammatical errors as well)

% @suvodm - add cadec model here
\section{Comparison to previous results} 

Comparisons to previous work are not straightforward due to the variability in training data used, number of model parameters, hyper-parameter tuning and other train-time parameters. 

Reported results as well as the replicated results below show that we compare favorably to previous work, despite not using language-specific tuning or techniques. The best EN-DE ContraPro accuracy using parallel data reported in \cite{DBLP:journals/corr/abs-2010-09482} is 0.83 on a model trained with 22.4M segments, which is 0.05 higher than our Ctx-Deep model, trained with 4M segments. On the same test set \cite{fernandes21acl} shows an accuracy score of 0.66, \cite{lopes-etal-2020-document} reports a maximum performance of 0.71 while \cite{maruf-etal-2019-selective} reports a maximum of 0.73/0.69 with an offline and an online model respectively.

% The best EN-DE ContraPro accuracy using parallel data reported in \cite{fernandes21acl} is 0.66, obtained with a concatenated input method. Among all the methods tested, this is only outperformed by a model that additionally performs pre-training on a large monolingual corpus of 82M segments (0.80 accuracy, 0.02 higher than our Ctx-Deep model). \gd{Add Number of parameters of this model} On the same test set \cite{lopes-etal-2020-document} reports a maximum performance of 0.71 while \cite{maruf-etal-2019-selective} reports a maximum of 0.73/0.69 with an offline and an online model respectively. \cite{DBLP:journals/corr/abs-2105-06977}

On EN-FR \cite{DBLP:journals/corr/abs-2105-06977} report an accuracy of 0.91, to our knowledge the best reported results on this test set. We reach 0.90/0.86 accuracy on this test with the Ctx-Deep and the student model respectively. On EN-RU \cite{voita-etal-2019-good} also obtains the best results when using concatenation models, however on average these results are lower than the Ctx-Deep ones. \cite{voita-etal-2019-context} introduces DocRepair, a two-pass method which obtains optimal results but has considerable computational drawbacks compared to our proposal. DocRepair is tested on EN-RU and achieves 0.92 on deixis, 0.75/0.86 on ellipsis and an impressive 0.80 on lexical cohesion.

% On EN-FR \cite{lopes-etal-2020-document} report an accuracy of 0.83, to our knowledge the best reported results on this test set. We reach 0.90/0.86 accuracy on this test with the Ctx-Deep and the student model respectively. On EN-RU \cite{voita-etal-2019-good} also obtains the best results when using concatenation models, however on average these results are lower than the Ctx-Deep ones. \cite{voita-etal-2019-context} introduces DocRepair, a two-pass method which obtains optimal results but has considerable computational drawbacks compared to our proposal. DocRepair is tested on EN-RU and achieves 0.92 on deixis, 0.75/0.86 on ellipsis and an impressive 0.80 on lexical cohesion.

%However we believe that our results compare favorably to previously results, either by reproducing results from previous studies or by tackling these languages individually. 

% cadec is two step process - and much more complex - while our is simple one step process.
We were able to perform a side-by-side comparison of our proposed approach with that of ~\cite{voita-etal-2019-good} on EN-RU, due to availability of the implementation. We reproduce the results of the \textbf{CADec} model by using the use same EN-RU train data set and configuration as in the original paper~\cite{voita-etal-2019-good}. The train data contains 7.5 million parallel segments, out of which 1.5 million are contextual. For comparison,  we train a contextual model with 6:6 encoder:decoder layers, totaling the same number of parameters as CADec. During testing, the source segment is translated using its preceding context and the context translation is subsequently stripped from the output. 

The results are shown in Table \ref{tab:compare_with_SOTA}. We report the results for both non-targeted and targeted test sets. Ctx-6:6 shows high quality, comparable to the state of the art CADec model on three of the test sets, and better in two of them (ellipsis-vp and ellipsis-infl). %A detailed language specific comparison can be found in the Appendix section~\ref{sec:app:results}.

\begin{table}[]
\small
\centering
{\small{
\begin{tabular}{lllc|cp{0.1\linewidth}}
Arc &	Metric&	Test set&	Targeted &	CADec &	Ctx-6:6 \\ \hline
RU &	BLEU& CADec test  &			& 30.1 & \textbf{30.4} \\	
	&Acc& Deixis &	\checkmark		& \textbf{0.81} & 0.80 \\
    &Acc& Lex-coh &	\checkmark		& 0.47 & \textbf{0.48} \\	
	&Acc& Ellipsis-vp &	\checkmark	& 0.69 & \textbf{0.73} \\
	&Acc& Ellipsis-infl &	\checkmark	& 0.56 & \textbf{0.75}\\\hline

\end{tabular}
}}
\caption{Comparison of concatenated contextual (Ctx-6:6) and Context Aware Decoder(CADec) models.}
\label{tab:compare_with_SOTA}
\end{table}

\section{Related work}
\label{sec:background}

\subsection{Document translation}
% \gd{can def. be shortened if needed}}

A straight-forward way to include context proposed by \citet{tiedemann-scherrer-2017-neural} is to train a standard NMT model on \textit{pseudo}-document parallel data obtained by concatenating two or more consecutive sentences. As large-scale document-level parallel data is not widely available, prior work explored data augmentation: augmenting the training data with back-translated monolingual documents \citep{junczys-dowmunt-2019-microsoft} and leveraging monolingual data to train a document-level \textit{repair} system \citep{voita-etal-2019-context} .
%Training on pseudo-documents can be improved by  filtering the context words (e.g. only named entities and rare words)\citep{kim-etal-2019-document}, augmenting the training data with back-translated monolingual documents \citep{junczys-dowmunt-2019-microsoft}, or by using a bilingual encoder that can encode both the source and target context \citep{berard-EtAl:2020:WMT}.

%Offline methods have also been proposed that take advantage of monolingual document-level data which, unlike parallel data, is available at large scale even for low resource languages. In \citep{voita-etal-2019-context} sentences are first translated in isolation, and then post-edited using a monolingual document-level \textit{repair} system. While showing very good performance, two-pass approaches have practical limitations: they require maintaining and deploying an additional model for each target language and they double the decoding time.

To scale to larger context beyond the previous or next sentence, prior work proposed changes to the architecture to improve how context is compressed and attended to by the encoder and/or decoder: multi-encoder architectures \citep{bawden-etal-2018-evaluating},  hierarchical or sparse attention over context sentences \citep{maruf-etal-2019-selective, miculicich-etal-2018-document, bao-etal-2021-g}, incorporating diverse pretrained context representations that combine local and document-level context \citep{DBLP:conf/iclr/ZhuXWHQZLL20, donato-etal-2021-diverse}. 
%selecting the most salient context for decoding the current sentence by adding a hierarchical ..
%Some work uses auxiliary losses to enforce consistency \citep{jwalapuram-etal-2020-pronoun, sperber-etal-2020-consistent}. 
However, recent work \citep{fernandes21acl} has shown scaling up to longer contexts brings diminishing returns or can even hurt performance due to instability in training attention and positional embeddings \citep{nguyen-etal-2021-data}.

\subsection{Large models and Knowledge distillation}

Prior work has showed that increasing model capacity in MT is beneficial particularly when increasing the size or the diversity of the training data, such as for multi-lingual models. With respect to depth versus width, \cite{DBLP:journals/corr/abs-2001-08361} do an extensive comparison of model capacity for neural language models and find that increasing width and depth while maintaining capacity gives similar results.

Knowledge distillation \cite{DBLP:journals/corr/HintonVD15} (KD) has been introduced as a way to compress larger models, or ensembles thereof, into smaller more computationally efficient models that reach similar performance. For sequence to sequence models, \cite{kim-rush-2016-sequence} introduced sequence-level knowledge distillation for improved MT. Subsequently knowledge distillation has proved beneficial for non-autoregressive MT. In general, distillation is thought to be effective in low data regimes, small data sets, domain adaptation, transfer learning (e.g. \citet{currey-etal-2020-distilling}) which makes it particularly suitable for document-level translation, where parallel data is a bottleneck. %(See \cite{DBLP:journals/corr/Lopez-PazBSV15} for a theoretical perspective).

A significant body of work has been devoted to understanding why distillation works, such as \citep{gordon2019explaining, xu-etal-2021-distilled, Zhou2020Understanding} among others. While our work does not focus on investigating why distillation works, we do contribute the observation that students prove to be very robust and out-perform teachers on out-of-distribution data when input length is considered. %\mn{TODO: add back results on length distribution} For future work we plan to further investigate this observation and \gd{... ? XX }

\section{Conclusion}
\label{sec:conclusion}
% add deeper vs wider here? @suvodm
In this paper we address the task of contextual translation by using multi-segment Transformer models. We show that we can successfully push the limits of this approach to achieve robust and high performance across several languages and benchmarks, without any language or task-specific tuning. This is achieved by training models to perform both contextual and single-segment translation, which has the added benefit of improving single-segment translation quality. We also show that -- with fixed data conditions and model capacity -- deeper models are superior to wider models in modeling contextual dependencies between pronouns and their antecedents.

Next we showed that the increased computational burden can be mitigated through distillation. Finally we observe that distilled models are more robust than their teachers on long input, which opens a new direction for improving MT models though distillation.

In this paper we have kept the training data fixed and have not investigated any data manipulations that could lead to improved performance. Particularly, our results indicate that standard document-level parallel data sets (such as the \textit{non-targeted} sets used in this paper) exhibit a limited amount of discourse phenomena. In this light, multi-segment models trained on similar data may not learn to pay sufficient ``attention" to context. In future work, we plan to investigate if parallel data can be improved by measuring and controlling context-sensitivity.
%Altering the training data distribution by detecting and up-sampling context-sensitive data is a promising direction to address this. 

%The current work focuses on a fixed architecture, the Transformer, and uses it to model longer sequences in training of MT models. In future work, we plan to investigate architectures optimized to process longer text input \cite{dai-etal-2019-transformer,rae2019compressive,beltagy2020longformer}. This is a scalable approach for improved contextual translation, taking advantage of architectures developed for efficient processing of long text input irrespective of the task. 

%In this work we showed that on three language pairs, En→De, Fr, Ru, increasing the capacity of a contextual model by adding more decoder layers improves targeted accuracy by 7pp on average (0.7→0.77, 0.73→ 0.84, 0.82→ 0.86) on discourse phenomena test sets, more so than increasing the width of feed-forward layers. Using knowledge distillation, we can improve the shallower, lower capacity model’s performance to a significant degree (3pp on average). Future research direction will address weakness of the contextual models wrt very long input sequences and error propagation.

\newpage
\section{Ethical Considerations}

In this work we used professional translators to evaluate the quality and accuracy of our models on publicly available data sets. The professional translators were recruited by a language service provider and were compensated according to industry standards. 

\section{Limitations}
Every approach has its limitations, and our approach is no exclusion to that. Our contextual model adopts a simple approach of concatenating the previous sentence to the current sentence. While this improves the contextual model's performance significantly, we have not experimented the effect of context size on model performance and have used standard context length from literature. 

Also, when comparing with previous results, we have reproduced a state of the art approach (CADec) on a standard data set for EN-RU arc. For the remaining language pairs and test sets, we have cited previously reported results without reproducing them.

%Finally, contextual model are often language dependent, often if the target language doesn't have gender specific pronouns or different formalities (i.e., English), then often contextual models will not have significant advantages over sentence level models. 
Finally, the targeted test sets used are limited wrt the different discourse phenomena they explore: anaphoric pronouns, lexical cohesion and verb forms. We have not investigated if our approach impacts other discourse phenomena or if it affects biases in translation. 

%Also some targeted test sets contains contrastive segments which either has intrasegmental dependencies or has limited discourse phenomenon. This results in minimal gain in performance between contextual and baseline models. 

% Finally, during inference time, our multi-segment input is simply translated as a continuous output sequence. This increases translation time, which can be crucial for time critical operations (i.e., live subtitle translations). 

% 1. required Context not adjacent to current sentence - then the context will not be useful. * may be 2to2 to nton to full2full
% 2. As we translate both context and current sentence - the inference time is much larger
% 3. use of context is language dependent. (specifically non-gender specific languages)
% 4. test sets doesn't contain too much discourse phenomenon
% 5. do not reinvestigate length of context - 
% 6. We did not investigate whether out model affects other discourse phenemenon like bias 

\bibliography{anthology,custom}
\bibliographystyle{acl_natbib}

\appendix

\clearpage

\section{Experimental Settings}
\label{sec:app:sockeye_arguments}

NMT models were built using the Transformer-base architecture \citep{vaswani2017attention}. The source embeddings, target embeddings, and the output layer's weight matrix are tied. Training is done on 8 GPUs with Sockeye 2's large batch training.
 
Parameters for training a 6:8 encoder:decoder model (same parameters are used for the other encoder:decoder configurations):
\begin{lstlisting}[language=dict, basicstyle=\tiny]
Arguments: Namespace(allow_missing_params=False, amp=False, amp_scale_interval=2000, batch_sentences_multiple_of=8, batch_size=8192, batch_type='word', bucket_scaling=False, bucket_width=8, cache_last_best_params=0, cache_metric='perplexity', cache_strategy='best', checkpoint_improvement_threshold=0.0, checkpoint_interval=4000, config=None, decode_and_evaluate=-1, decode_and_evaluate_device_id=None, decoder='transformer', device_ids=[-8], disable_device_locking=False, dry_run=False, dtype='float32', embed_dropout=(0.0, 0.0), encoder='transformer', env=None, fixed_param_names=[], fixed_param_strategy=None, gradient_clipping_threshold=1.0, gradient_clipping_type='none', horovod=False, ignore_extra_params=False, initial_learning_rate=0.0002, keep_initializations=False, keep_last_params=60, kvstore='device', label_smoothing=0.1, learning_rate_reduce_factor=0.9, learning_rate_reduce_num_not_improved=8, learning_rate_scheduler_type='plateau-reduce', learning_rate_t_scale=1.0, learning_rate_warmup=0, length_task=None, length_task_layers=1, length_task_weight=1.0, lhuc=None, lock_dir='/tmp', loglevel='INFO', loglevel_secondary_workers='INFO', loss='cross-entropy-without-softmax-output', max_checkpoints=None, max_num_checkpoint_not_improved=30, max_num_epochs=None, max_samples=None, max_seconds=1036800, max_seq_len=(200, 200), max_updates=None, min_num_epochs=1, min_samples=None, min_updates=None, momentum=None, monitor_pattern=None, monitor_stat_func='mx_default', no_bucket_scaling=None, no_bucketing=False, no_hybridization=False, no_logfile=False, num_embed=(None, None), num_layers=(6, 8), num_words=(0, 0), omp_num_threads=None, optimized_metric='perplexity', optimizer='adam', optimizer_params=None, output='deep68_6_8', overwrite_output=True, pad_vocab_to_multiple_of=None, params=None, prepared_data='../data_en_de', quiet=False, quiet_secondary_workers=False, round_batch_sizes_to_multiple_of=None, seed=1, shared_vocab=False, source=None, source_factor_vocabs=[], source_factors=[], source_factors_combine=[], source_factors_num_embed=[], source_factors_share_embedding=[], source_factors_use_source_vocab=[], source_vocab=None, stop_training_on_decoder_failure=False, target=None, target_factor_vocabs=[], target_factors=[], target_factors_combine=[], target_factors_num_embed=[], target_factors_share_embedding=[], target_factors_use_target_vocab=[], target_factors_weight=[1.0], target_vocab=None, transformer_activation_type=('relu', 'relu'), transformer_attention_heads=(8, 8), transformer_dropout_act=(0.1, 0.1), transformer_dropout_attention=(0.1, 0.1), transformer_dropout_prepost=(0.1, 0.1), transformer_feed_forward_num_hidden=(2048, 2048), transformer_feed_forward_use_glu=False, transformer_model_size=(512, 512), transformer_positional_embedding_type='fixed', transformer_postprocess=('dr', 'dr'), transformer_preprocess=('n', 'n'), update_interval=1, use_cpu=False, validation_source='../data_en_de/', validation_source_factors=[], validation_target='../data_en_de/', validation_target_factors=[], weight_decay=0.0, weight_init='xavier', weight_init_scale=3.0, weight_init_xavier_factor_type='avg', weight_init_xavier_rand_type='uniform', weight_tying_type='src_trg_softmax', word_min_count=(1, 1))
\end{lstlisting}

All data except Chinese(ZH) is tokenized using the Python Moses tokenizer at \url{https://github.com/alvations/sacremoses}. Chinese is tokenized using Jieba tokenizer at \url{https://github.com/fxsjy/jieba}. Words were segmented using BPE \citep{sennrich-etal-2016-neural} with 32K operations. Source and target subwords shared the same vocabulary and training segments longer than 95 tokens were removed.

% updated with new lang pair @suvodm 
%TODO: Add new lang details
\begin{table}[h]
\centering
\begin{tabular}{lll}
\hline
\textbf{LP} &  \textbf{\#segments} & \textbf{Data source}\\
%\textbf{LP} &  \textbf{Nb} & \textbf{Data} & \textbf{Phenomenon} & \textbf{Nb} & \textbf{Prior accuracy} & \textbf{Data}\\
% &  \textbf{Segments} & \textbf{source} & & \textbf{Segments} & \textbf{reported} & \textbf{source}\\
\hline
EN-DE & 4M & WMT, 2019 \\
EN-FR & 9.8M & Open Subtitles 2018\\
EN-RU & 8.7M & Open Subtitles 2018\\
ZH-EN & 17.4M & UN Parallel Corpus V1.0\\\hline
\end{tabular}
\caption{Document parallel training data.} %Multi-segment models are trained on twice the amount of data: segments in isolation, as well as concatenated with the preceding segments.}
\label{tab:Datasets}
\end{table}

\newpage
\section{Multi-segment translation models}
\label{sec:app:Multi-segment translation models}
Our contextual model is a multi-segment model. The document-level parallel data is transformed to contain concatenated, multi-segment input for our models. Specifically, we concatenate two consecutive segments, both source and target, to create a new data point. training a model with just concatenated inputs will render the model useless for translating isolated segments, while training a model with just isolated segments will not be useful for contextual translations. Thus we aim aim to create a single translation models which can perform both translation in-context and in isolation. We do this by duplicating the training data by concatenating in-context data with isolated data. See Algorithm~\ref{algo:1} for the pseudo-code for this concatenation process. 

\begin{algorithm}[h]
    \SetKwInOut{Input}{Input}
    \SetKwInOut{Output}{Output}
    
    \SetKwData{T}{T}
    \SetKwData{Con}{{Concat}}
    \SetKwData{doc}{{Doc}}
    \SetKwData{newDoc}{Doc} 
    
    \Input{Parallel train data $\T$}
    \Output{Augmented parallel train data}
    
    $\T_{ctx} \leftarrow []$
    
    \For{$\doc^s,\doc^t \in \T$}
    {
        $\newDoc_{ctx}^s \leftarrow []$
        
        $\newDoc_{ctx}^t \leftarrow []$
        
        \For{$i \in 0 \dots len(\doc)-1$}
        {
            $\newDoc_{ctx}^s \leftarrow \newDoc_{ctx}^s + \Con(\doc^s_{i},\doc^s_{i+1})$
            
            $\newDoc_{ctx}^t \leftarrow \newDoc_{ctx}^t + \Con(\doc^t_{i},\doc^t_{i+1})$
            
        }
        
        $\T_{ctx} \leftarrow \T_{ctx} + \newDoc_{ctx}$
        
    }
    
    \KwRet{$T + \T_{ctx}$}
    \caption{Algorithm to transform the training data.}
\label{algo:1}
\end{algorithm}

% @suvodm add previous results for ZH EN arc
\section{Results}
\label{sec:app:results}
\paragraph{Comparison to previous results} The best EN-DE ContraPro accuracy using parallel data reported in \cite{fernandes21acl} is 0.66, obtained with a concatenated input method. Among all the methods tested, this is only outperformed by a model that additionally performs pre-training on a large monolingual corpus (0.80 accuracy, 0.02 higher than our Ctx-Deep model). On the same test set \cite{lopes-etal-2020-document} reports a maximum performance of 0.71 while \cite{maruf-etal-2019-selective} reports a maximum of 0.73/0.69 with an offline and an online model respectively.

On EN-FR \cite{lopes-etal-2020-document} report an accuracy of 0.83, to our knowledge the best reported results on this test set. We reach 0.90/0.86 accuracy on this test with the Ctx-Deep and the student model respectively. On EN-RU \cite{voita-etal-2019-good} also obtains the best results when using concatenation models, however on average these results are lower than the Ctx-Deep ones. \cite{voita-etal-2019-context} introduces DocRepair, a two-pass method which obtains optimal results but has considerable computational drawbacks compared to our proposal. DocRepair is tested on EN-RU and achieves 0.92 on deixis, 0.75/0.86 on ellipsis and an impressive 0.80 on lexical cohesion.

% # Update table with baseline, student model
\begin{table}[]
\centering
{\small{
\begin{tabular}{lp{0.15\linewidth}p{0.15\linewidth}p{0.15\linewidth}p{0.15\linewidth}}
\hline
\textbf{Model} &  \textbf{Enc. blocks} & \textbf{Dec. blocks} & \textbf{FF \ size} & \textbf{Total param.}\\
\hline
Ctx & 6 & 2 & 2048 & 44 M \\
\hline
Deep-52 & 6 & 4 & 2048 & 52 M \\
Wide-52 & 6 & 2 & 3072 &  \\
\hline
Deep-60 & 6 & 6 & 2048 & 60 M \\
Wide-60 & 6 & 2 & 4096 &  \\
\hline
Deep-68 & 6 & 8 & 2048 & 68 M \\
Wide-68 & 6 & 2 & 5120 &  \\
\hline
Deep-76 & 6 & 10 & 2048 & 76 M \\
Wide-76 & 6 & 2 & 6144 &  \\
\hline
\end{tabular}
}}
\caption{Multi-segment models. For each increase in model capacity, measured in millions or parameters, we vary the width of the FF layer and the depth of the decoder, to obtain a deep and a wide configuration.}
\label{tab:ModelCapacityAppendix}
\end{table}

\subsection{Segment Length Analysis}
\label{sec:app:Segmentlength}

\textbf{Non-targeted} test sets shows the  best  performance is obtained by the deep multi-segment models used \textit{without context} in Table~\ref{tab:results5}. The result shows significant difference in performance between translation done with and without context in case of English (EN) to Russian (RU) arc.
 
We analyze the  non-targeted test set with respect to segment length of the training data. We compute the median segment length of the training data is 15 with a sd of 9. We further split the test set in 4 partitions with respect to the relation to this training data median and sd. Table~\ref{tab:legthAnalysisAppendix} shows the performance of each partition when translation is done with and without context. We also see when the test segment length deviates by more than 4 sd, the performance of translation with context degrades significantly and $\approx 75\%$ segments falls into this partition.  We think this length deviation between training and test set is the reason behind the performance drop between with and without context translations.

\begin{table}[]
\centering
\small
\begin{tabular}{l|l|ll}
\multirow{2}{*}{Len. Range} & \multicolumn{1}{c|}{\multirow{2}{*}{\#Segments}} & \multicolumn{2}{c}{BLEU}                                                                                                                                     \\
                            & \multicolumn{1}{c|}{}                            & \multicolumn{1}{c}{\begin{tabular}[c]{@{}c@{}}Without \\ Context\end{tabular}} & \multicolumn{1}{c}{\begin{tabular}[c]{@{}c@{}}With \\ Context\end{tabular}} \\ \hline
(0,m)               & 22                                               & 5.7                                                                            & 6.5                                                                         \\
(m,m + 2* SD]                   & 351                                              & 17.3                                                                           & 19.1                                                                        \\
(m+2*SD, m+4 *SD]                  & 652                                              & 20.3                                                                           & 21.5                                                                        \\
(m+4* SD,$+\infty$)            & 3077                                             & 19.6                                                                           & 16.7                                                                        \\ \hline
\end{tabular}
\caption{BLEU scores on WMT20 EN-RU test set. The test set is split into four partitions, wrt. the input length. For example, the Med - 2sd bin contains input of length between the median length seen in training and 2 standard deviations.}
\label{tab:legthAnalysisAppendix}
\end{table}

\end{document}